\documentclass[10pt,twocolumn]{article}
\usepackage[letterpaper,margin=0.75in]{geometry}
\usepackage{cite}
\usepackage{amsmath,amssymb,amsfonts}
\usepackage{algorithm}
\usepackage{algpseudocode}
\usepackage{graphicx}
\usepackage{textcomp}
\usepackage{import}
\usepackage{gensymb}
\usepackage{comment}

\begin{document}

\title{\bf Standardisation of Convex Ultrasound Data Through Geometric Analysis and Augmentation}

\author{%
Alistair Weld\thanks{The Hamlyn Centre, Imperial College London, London, UK. (e-mail: a.weld20@imperial.ac.uk).} , 
Giovanni Faoro\thanks{The BioRobotics Institute, Scuola Superiore Sant’Anna, 56127 Pisa, Italy.} ,
Luke Dixon\thanks{Department of Imaging, Charing Cross Hospital, Fulham Palace Rd, London, W6 8RF, UK.} , \\
Sophie Camp\thanks{Department of Neurosurgery, Charing Cross Hospital, Fulham Palace Rd, London, W6 8RF, UK.},
Arianna Menciassi\footnotemark[2] , 
Stamatia Giannarou\footnotemark[1] , 
}

\maketitle

\begin{abstract}
The application of ultrasound in healthcare has seen increased diversity and importance. Unlike other medical imaging modalities, ultrasound research and development has historically lagged, particularly in the case of applications with data-driven algorithms. A significant issue with ultrasound is the extreme variability of the images, due to the number of different machines available and the possible combination of parameter settings. One outcome of this is the 
lack of standardised and benchmarking ultrasound datasets. The method proposed in this article is an approach to alleviating this issue of disorganisation. For this purpose, the issue of ultrasound data sparsity is examined and a novel perspective, approach, and solution is proposed; involving the extraction of the underlying ultrasound plane within the image and representing it using annulus sector geometry. An application of this methodology is proposed, which is the extraction of scan lines and the linearisation of convex planes. Validation of the robustness of the proposed method is performed on both private and public data. The impact of deformation and the invertibility of augmentation using the estimated annulus sector parameters is also studied. Keywords: Ultrasound, Annulus Sector, Augmentation, Linearisation.
\end{abstract}

\section{Introduction}

The role of ultrasound (US) in healthcare has increased greatly over the last three decades and is now the main imaging modality in the evaluation of many different organ systems and pathologies. The ability of US to provide dynamic imaging in real time, combined with its relative affordability and adaptability to various clinical settings, has solidified its position as a useful tool in US-guided procedures and surgical guidance \cite{Dixon2022IntraoperativeUI} \cite{Krekel2013IntraoperativeUG} \cite{Lubner2021DiagnosticAP}. As there is a global increase in demand for medical imaging (in all modalities) coupled with a global shortage of experts \cite{Won2024SoundTA}, the research on technical solutions has also increased considerably, notably within the space of data-driven algorithms \cite{Jabeen2022BreastCC} \cite{Roy2020DeepLF} \cite{Akkus2019ASO}, robotics and control \cite{Jiang2023RoboticUI}.

\begin{figure}[ht]
\centering
\frame{\includegraphics[width=0.9\columnwidth]{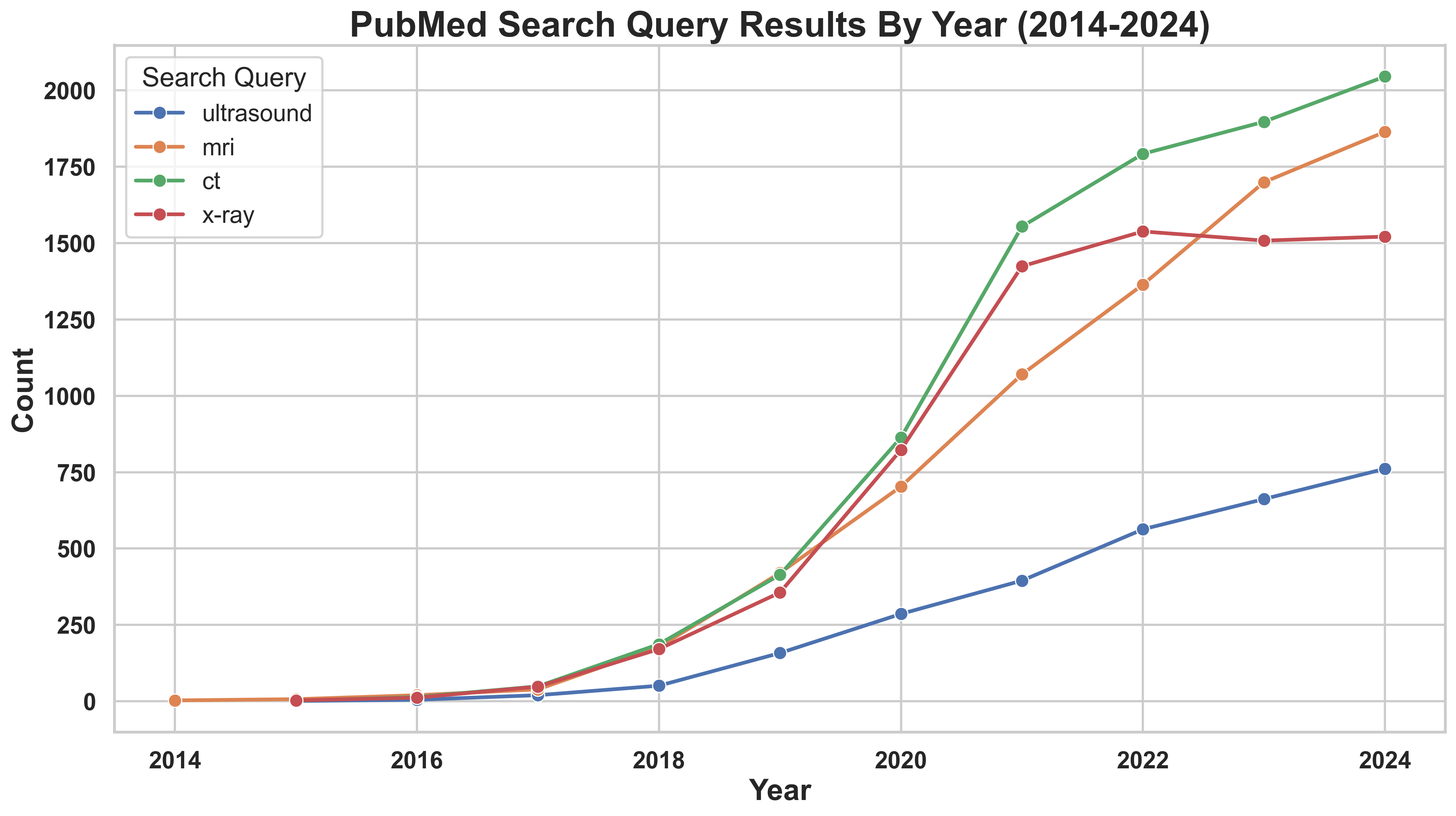}}
\caption{Medical imaging modality deep learning research output trends.}
\label{fig:search}
\end{figure}

Unlike other imaging modalities such as magnetic resonance imaging (MRI) \cite{Menze2015TheMB}, public US data is relatively disorganised and sparse; lacking standardised and benchmarking datasets. In particular, for data-driven methods, this has presented a bottleneck in relevant research. This is a significant factor that has resulted in fewer US research output, relative to other modalities. For example, a simple search of \texttt{https://pubmed.ncbi.nlm.nih.gov} using the keyword queries "modality" AND "deep learning" shows that US lags far behind other modalities, as is shown in Fig.~\ref{fig:search}. US images are highly variable depending on the machine used, the wide range of different probes, machine settings, capture methods, and operator dependence. The slight alterations in these variables may account for substantial differences in the US plane and image intensities. Since the publication of US data is often targeted at clinical audiences (e.g. \texttt{ https://radiopaedia.org}), or consists of data collected by a single centre, there is no common consensus on standardising how to publish US data. This often leads to publishing images with manual modifications, such as cropping or the addition of additional text or graphical features.  Particularly for convex US images, the variable curved geometry and decoupling of image height and scan line depth create additional data complexity and variability. These variation factors compound the difficulty of using public data for research and development. 

From a technical perspective, this lack of standardisation can be perceived as equivalent to a lack of data. This has consequences for data-driven methods that either train on large volumes of data or that need to be evaluated on diverse datasets to prove robustness and generalisability. The option of manually collecting and annotating data is labour intensive and involves logistical variables such as ethics and data sharing between multiple centres. For example, the well-known breast US dataset BUSI \cite{AlDhabyani2019DatasetOB}, which only contains 780 images, took about a year to collect and annotate. Because of these reasons, synthetic phantoms are often substituted for in vivo data, at the cost of data realism. Or, research is published using private datasets exclusively, making reproducibility an issue. Technical methods for alleviating the issue of data scarcity and lack of standardisation are limited. The main solution proposed is the application of generative models for the generation of synthetic medical datasets \cite{Pinaya2022BrainIG} or for embedding into training frameworks where real images or annotations are limited or perhaps not available. However, there are many technical and ethical concerns and complications when it comes to the use of synthetic data in clinical applications \cite{Susser2024SyntheticHD} \cite{Giuffr2023HarnessingTP} \cite{Singh2024IsSD}. 

In this work, an approach to standardise US datasets is proposed based on automatically extracting the US plane from convex US images. 
Simple use cases include normalising the position of the US plane within a 2D array to more advanced applications such as data augmentation for neural network pipelines \cite{Chen2024COVIDNetLA} or signal processing methods that process along scan lines \cite{Karamalis2012UltrasoundCM}. 
From a linear algebra perspective, performing matrix operations, such as convolutions or matrix multiplications, becomes more intuitive and logical when scan lines align with image columns. Therefore, having the convex plane extracted enables adjustment of these operations or the linearisation of the plane. A key limitation of these approaches is their reliance on the convex US plane being known a priori, where in the above two references manual labelling of the plane was done.

To our knowledge, no other work has focused on the specific task of automatically extracting the convex US plane. Methods have been explored to minimise manual annotation and data organisation labour. For example, determining the orientation and location of US scans relative to anatomical structures has been proposed to support diagnosis. These works often focus on the identification of the foetal standard plane \cite{Pu2021AutomaticFU} \cite{Baumgartner2016SonoNetRD}. Other works include partitioning of data sets based on the view of the lung \cite{VanBerlo2022EnhancingAE}. Additional topics include the automation of US quality assessment, which helps to ensure correct biological measurements \cite{Wu2017FUIQAFU}, and could be also used for data set organisation. However, the limitations of these methods are the lack of explicit isolation of the relevant, full US-plane geometry and the limited use-case to specific anatomy or US-task. Furthermore, most proposed methods are based on deep learning, which introduces classic black-box interpretability and reliability concerns.

In this paper, we introduce a nonanatomy- or application-specific method for automatic extraction of the visible, convex US plane. This is done using traditional image processing techniques and representing the convex plane using annulus sector geometry. Our key contributions are as follows:

1) We propose the first framework for automatically extracting convex US planes from nonstandardised US images, including those affected by common GUI elements, additional text and graphical features, and plane modifications such as cropping.

2) We demonstrate that convex US planes can be effectively represented using annulus sector geometry, enabling transformations that preserve the geometry of underlying features.

The first evaluation conducted is the accuracy and robustness of the proposed method in estimating the annulus sector parameters. This included an evaluation using both public and private datasets, where the ground-truth annotations were created manually. The data used for the evaluation included cases where there was incomplete visualisation of the convex plane and corruption. Further evaluation explored the viability of using the method to augment the image, including the invertibility of the process and measuring the deformation of a circular object post-augmentation.

\section{Methodology}

\begin{figure}[t]
\centering
\frame{\includegraphics[width=0.9\columnwidth]{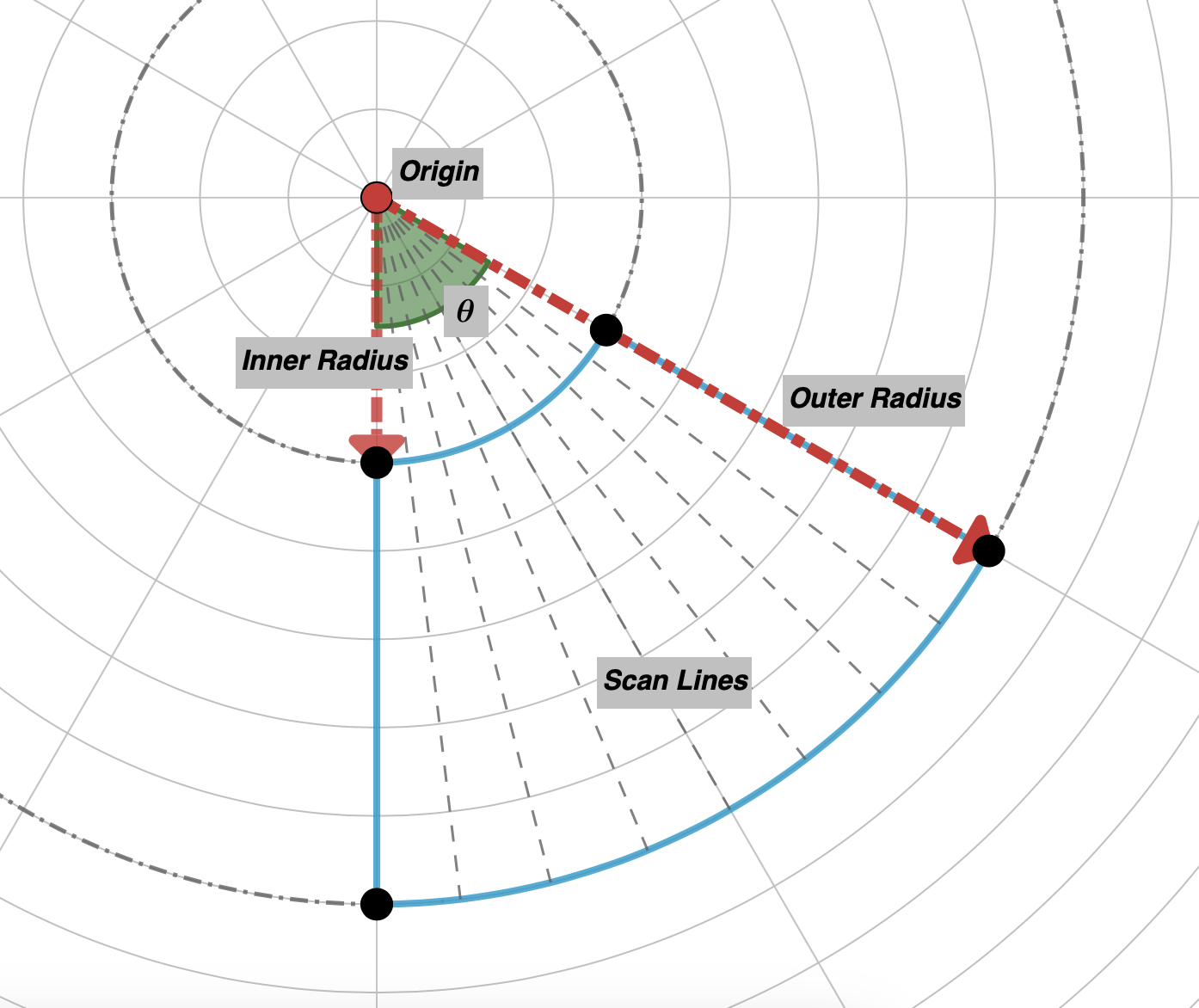}}
\caption{Annulus sector geometric properties, created using \texttt{https://www.geogebra.org}.}
\label{fig:ccl}
\end{figure}

In this section, a description is given of the method for extracting the annulus sector parameters, highlighted in Fig.~\ref{fig:ccl} from convex US images, e.g. left subfigure in Fig.~\ref{fig:meth1}. The proposed method can be broken down into four stages: 1) "Ultrasound Plane Masking", which aims to find the rough US plane and represent it as a binary mask, and to also remove the background GUI; 2) "Centre Of Plane Calculation" involves determining the centre line through the US plane; 3) "Radial Boundary Detection" estimating the radial boundaries of the plane, utilising the symmetry from the centre line; 4) "Annulus Sector Parameter Calculation", is the final stage of the algorithm where the annulus sector parameters are determined using the information from the previous stages. At the end of this section, the methodology for how the annulus sector parameters can be used for the automatic linearisation of a convex US plane is described. When required, we simplify the mathematical description with algorithmic notation, e.g. pythonic indexing or for loops such as [0] and [-1] for indexing the first and last element in a list, respectively. 


\subsection{Ultrasound Plane Masking}

\begin{figure}[ht]
\centering
\includegraphics[width=\columnwidth]{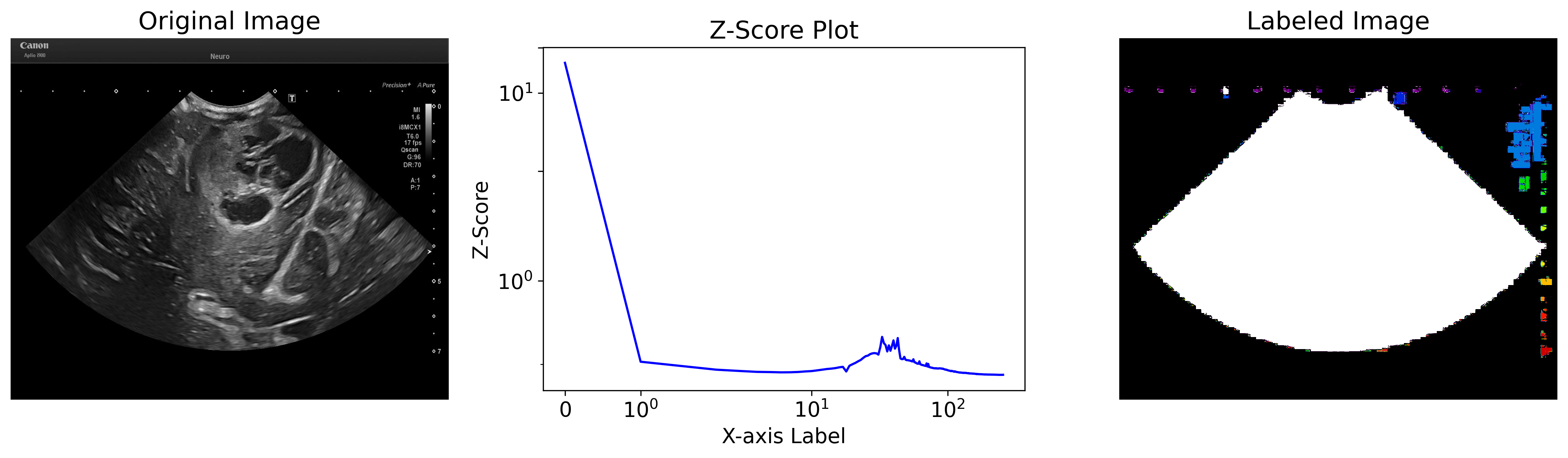}
\caption{Extracting the US plane. Left shows the original image. Middle shows a plot of the z-score, where in this circumstance there is only one spike as the background pixel is constant at 0. Right shows the output of the connected components process, where the US plane is the largest object and represented in white (the other colours represent other objects, belonging to the GUI).}
\label{fig:meth1}
\end{figure}

A raw 2D B-mode convex US image is defined as $I_{rgb}\in\mathbb{R}^{h,w,c}$ - an example of a standard US image with full GUI is shown in Fig.~\ref{fig:meth1}. This image is converted into grayscale $I_g$. We define the image origin at the top left corner. The superimposed GUI on the image, for example, the header and text relating to machine and parameter settings, should be removed because they may affect further image processing. To identify the US plane within the image, a simple two-pass 4-connectivity connected component labelling method \cite{He2017TheCL}, \cite{Ballard1982ComputerV} is used. First, the image is binarised to separate the foreground and black background pixels. Typically, the pixel intensity in the background, the area outside of the US plane, is equal to 0. However, this is not always the case due to variability on the data capture and storage. So, the background is detected by analysing the image's pixel intensity distribution. 

\begin{equation}
    H = hist(I_g, bins=[0,255]) 
    \label{eq:hist}
\end{equation}

As the background is homogeneous, its corresponding pixel intensities create a large spike within the intensity distribution. A spike in the histogram is determined if the frequency of an intensity is greater than an adaptive minimum threshold and is at least half the value of the highest bin frequency. This threshold ensures that only bins with sufficiently high frequencies relative to the peak frequency are considered. The z-score analysis is used for spike detection. Since background pixels are homogeneous, they produce a distinct spike, defined $\textit{}{sp_{thr}}$, in the histogram that meets our frequency threshold, regardless of high-intensity white pixels. We define the background intensity range using the furthest spike from the origin that satisfies the given criteria. Where the range is important to accommodate noise in the data capture or the saving. 
Determining the upper intensity of the background $bg_{max}$ enables the creation of the foreground-background mask as:

\begin{equation}
    Z[i]=\frac{H[i]-\mu}{\sigma}, \forall i \in [0..255]
    \label{eq:spike}
\end{equation}

\begin{equation}
\begin{aligned}
sp_{thr} &= \max\left(\frac{\max(Z)}{2},\, 2\right), \\
bg_{\max} &= \max\{\, i \in \{0,1,2,\ldots,255\} : Z_i > sp_{thr} \}.
\label{eq:spike_thresh}
\end{aligned}
\end{equation}

\begin{equation}
I_b = 
\begin{cases}
1 & \text{if } I_{g} > bg_{max}\\
0,              & \text{otherwise}
\end{cases}
\end{equation}
where, $\mu$ and $\sigma$ are the mean and standard deviation of the image's intensity distribution. Once the foreground-background mask $I_b$ has been constructed, a connected component function can identify distinct objects by grouping adjacent pixels, separating the US plane from other foreground components in the image (e.g text, GUI components):
\begin{equation}
P[j,i] = 
\begin{cases}
C_c & \text{if } f_{ccl}(I_b[j,i]) = C_c\\
0,              & \text{otherwise}
\end{cases}
\end{equation}
where, $f_{ccl}()$ denotes the two-pass, 4-connectivity connected component mapping. $C_c$ denotes the connected components index where $C = \{C_1, C_2, \ldots, C_n\}$, and $P$ is the output map of all connected components, that has the same dimensions as $I_b$. $f_{ccl}()$ works by assigning a pixel to a connected component $C_n$, if that pixel is contiguous either vertically or horizontally, to a pixel already belonging to $C_n$.

The convex US image plane can then be isolated through element-wise multiplication with the largest connected component subset as:
\begin{equation}
I_c = I_b \odot \text{argmax}_i(|C_i|), i \in \{1, \ldots, n\}
\end{equation}
where, $I_c$ denotes the image containing only the US plane. An overview of the extraction of the US plane is shown with an example in Fig.~\ref{fig:meth1}.

\subsection{Centre Of Plane Calculation}\label{sec:copc}

\begin{figure}[t!]
\centering
\includegraphics[width=0.65\columnwidth]{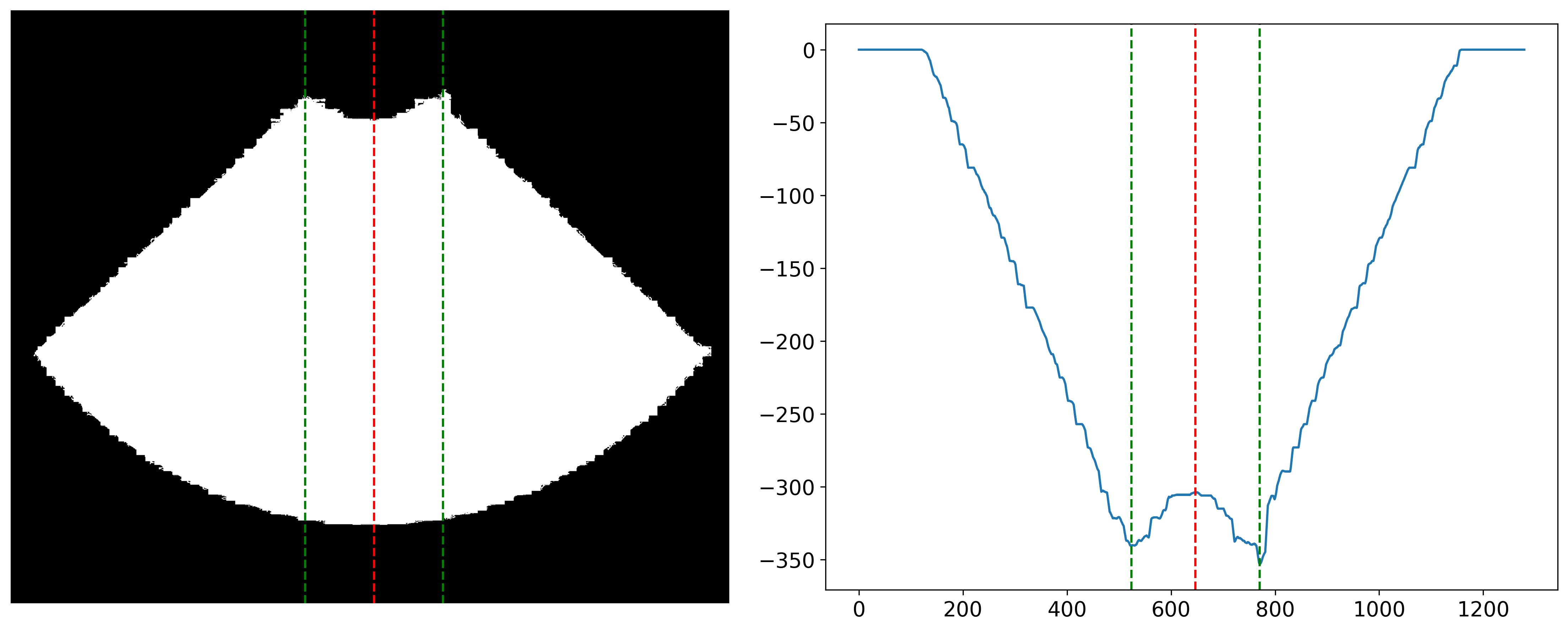}
\caption{(Right) An example of the calculation of the vertical axis of symmetry using the sliding window method. The blue line represents the intensities of vector $S$, the red line is the vertical line passing through point $m$ and the green lines are the vertical lines through points $min_l, min_r$. (Left) The binarised US plane with the same lines superimposed.}
\label{fig:meth2}
\end{figure}


The convex US image plane detected above can be represented as an annulus sector. The aim of this step is to use certain geometric assumptions to determine the parameters of this annulus sector such as the vertical axis of symmetry as shown in Fig.~\ref{fig:meth2}. The key to ensuring robustness of the representation of the US plane is to determine the true edges of the annulus sector, which may be affected by noise and artefacts. Assuming increasing signal attenuation with penetration depth and near-field echogenicity, the top corners of the annulus sector and the centre point of the inner arc, shown in Fig. \ref{fig:ccl}, are calculated first. To determine these 3 key points, a horizontal window $\hat{S}$ (with a height of one pixel and a width equal to that of the image) is incrementally slid over the image, moving only along the vertical direction, from the top of the image, to detect the inner arc.
\begin{gather}
\hat{S}^0 = [0]_{1 \times w} \nonumber \\
\hat{S}^j = \hat{S}^{j-1} - I_{c}[j,:], j=[1,\ldots, h/2], \hat{S} \leq 0
\end{gather}
\begin{equation}
S = \hat{S}^{h/2}*[0.2, 0.2, 0.2, 0.2, 0.2]
\end{equation}
where, $w$ and $h$ are the width and height of $I_c$, respectively and $*$ denotes convolution. The range of half the height has been arbitrarily chosen. 
The vector $S$ represents the accumulated row difference and is shown with the blue line in the right subfigure of Fig.~\ref{fig:meth2}. 

The centre of mass of the US plane is calculated to find an approximate of the centre of the inner curve. This is an intermediate step to prevent a negative impact in case there is asymmetry in the masked image $I_c$, which is likely due to image noise or superimposed GUI. An initial estimate of the $x$-coordinate of the centre of mass, $\hat{m}$, is calculated using the weighted average trapezoid rule as:
\begin{equation}
    \hat{m} = \frac{\sum^{w-1}_{i=1}\frac{iS[i]+(i+1)S[i+1]}{2}}{\sum^{w-1}_{i=1}\frac{S[i]+S[i+1]}{2}}
\end{equation}

Once $\hat{m}$ has been determined, the central convex hull is identified by locating the minima on either side (horizontal) of the centre of mass (displayed using the green vertical bars in Fig.~\ref{fig:meth2}). 
\begin{gather}
min_l=\text{min}(S[i]: i = 1..\hat{m}), \nonumber \\
min_r=\text{min}(S[i]: i = \hat{m}+1..w)
\end{gather}

To account for noise that may affect the mask, which would impact the convex hull, the final centre-of-mass estimation is calculated by averaging the centre of mass with the midpoint of $min_l$ and $min_r$.   

\begin{equation}
m=\frac{\hat{m}+(0.5(min_r-min_l)+min_l)}{2}
\end{equation}

Assuming that the US plane is normally orientated (without rotation), which is standard for 2D US, the coordinates of the inner and outer arc can be calculated by finding the first and final point of intersection of the US plane in $I_c$, along the vertical line passing through point $m$ as:
\begin{equation}
\begin{aligned}
S(m) &= \{\, j \in \{0,1,\ldots,h-1\} \mid I_c(j,m)=1 \,\}, \\[1mm]
\text{arc}_{\text{inner}} &= \bigl(\min S(m),\, m\bigr), \quad
\text{arc}_{\text{outer}} = \bigl(\max S(m),\, m\bigr).
\end{aligned}
\end{equation}

\subsection{Radial Boundary Detection}

The radial boundaries, i.e the legs of the convex plane of $I_c$, can be initially estimated using the farthest points, to the left and right of the $x$-coordinate of the centre of mass $m$:
\begin{equation}
\begin{aligned}
\hat{\text{edges}}_l &= \Biggl\{ \Bigl( j,\; \min\Bigl\{ i \in \{0,\dots,m-1\} : I_c(j,i)=1 \Bigr\} \Bigr) :\\[1mm]
                   &\quad j \in \{0,\dots,h-1\} \Biggr\},\\[1mm]
\hat{\text{edges}}_r &= \Biggl\{ \Bigl( j,\; \max\Bigl\{ i \in \{0,\dots,m-1\} : I_c(j,i)=1 \Bigr\} \Bigr) :\\[1mm]
                   &\quad j \in \{0,\dots,h-1\} \Biggr\}.
\end{aligned}
\end{equation}

The angular vertices that connect the radial boundaries and the outer arc are determined to find the end coordinates of the radial boundary. By determining the vertices, the edge points associated with the radial boundary are isolated and can be used to represent a prior estimate of the radial boundaries, although affected by noise and discontinuities along the US plane.
\begin{gather}
    [j_l,i_l] = \min_i(\{[j,i] \,|\, [j,i] \in vertices(hull(\hat{edges_{l}}))\}), \nonumber \\ 
    [j_r,i_r] = \max_i(\{[j,i] \,|\, [j,i] \in vertices(hull(\hat{edges_{r}}))\})
\end{gather}
\begin{gather}     
    edges_l = \{[j,i] \,|\, [j,i] \in \hat{edges_{l}} \wedge j \geq j_l\}, \nonumber \\     
    edges_r = \{[j,i] \,|\, [j,i] \in \hat{edges_{r}} \wedge j \geq j_r\} 
\end{gather}

To refine the estimation of the radial boundaries, RANSAC \cite{Fischler1981RandomSC} is applied to robustly regress a line of best fit, to determine the true radial boundaries of the plane. The RANSAC fit is performed for both the left $edges_l$ and right $edges_r$ set of edge points.
\begin{gather}     
[M_{l}, residual_l] = \text{RANSAC}_\text{fit}(edges_l) \nonumber \\   
[M_{r}, residual_r] = \text{RANSAC}_\text{fit}(edges_r) 
\end{gather} 
Here, $M$ are the model parameters and $residual$ is the fitting error. From the fitting, a linear representation of the radial boundaries is achieved as follows:
\begin{gather}     
\hat{RB_l} = \{M_{l}(min(edges_{l,i}), M_{l}(max(edges_{l,i})\}\nonumber \\   
\hat{RB_r} = \{M_{r}(min(edges_{r,i}), M_{r}(max(edges_{r,i})\}
\end{gather}

Assuming that the centre of mass $m$ splits the US plane symmetrically, the $\text{RANSAC}_\text{fit}$ producing the smallest residual error $[residual_l, residual_r]$ is chosen and re-projected across the vertical axis of symmetry defined by the line passing through $m$. This approach is based on the assumption that one side is likely to be worse effected by noise and discontinuities than the other. This is due to factors such as acoustic shadowing, which may result from orientation-related probe-tissue contact breaking. Using symmetry, the better-fit line is expected to be more reliable.
\begin{gather}
    RB_l = \begin{cases}
    \Phi - (\hat{RB_r} - \Phi), & \text{if } residual_r < residual_l \\
    \hat{RB_l}, & \text{otherwise }
    \end{cases}
\end{gather}

\begin{gather}
    RB_r = \begin{cases}
    \Phi + (\Phi - \hat{RB_l}), & \text{if } residual_r > residual_l \\
    \hat{RB_r}, & \text{otherwise }
    \end{cases}
\end{gather}
where, $\Phi = \{inner_{arc}, outer_{arc}\}$. 


\subsection{Annulus Sector Parameters Calculation}

The origin of the annulus sector can be determined by finding the point of intersection of the radial boundaries. The end points of the radial boundaries are defined as $RB_{l,top}$, $RB_{l,bottom}$, $RB_{r,top}$, $RB_{r,bottom}$. The origin $O$ is calculated by determining the coefficients $a,b,c$ of the line equation, followed by the Cramer's rule: 
\begin{gather}
    a = RB_{top}[1] - RB_{bottom}[1], \nonumber \\
    b = RB_{bottom}[0] - RB_{top}[0], \nonumber \\
    c = RB_{top}[0] \cdot RB_{bottom}[1] - RB_{bottom}[0] \cdot RB_{top}[1] \nonumber \\ 
    L = [a, b, -c]
\end{gather}

\begin{gather}
    D = L_l[0] \cdot L_r[1] - L_l[1] \cdot L_r[0], \nonumber \\
    Dx = L_l[2] \cdot L_r[1] - L_l[1] \cdot L_r[2], \nonumber \\
    Dy = L_l[0] \cdot L_r[2] - L_l[2] \cdot L_r[0], \nonumber \\
    O = \left[\frac{Dx}{D}, \frac{Dy}{D}\right]
\end{gather}

Once the origin $O$ has been estimated, the angle $\theta$ between the intersecting lines can then be calculated as:
\begin{equation*}
    \theta = \arccos{\frac{\overrightarrow{BA} \cdot \overrightarrow{BC}}{\|\overrightarrow{BA}\| \|\overrightarrow{BC}\|}}
\end{equation*}
where, $A=RB_{l,bottom}-O$, $B=O$, $C=RB_{R,bottom}-O$.



From this, the inner and outer radius of the annulus sector can be estimated using the point of origin and the point of intersection of the inner and outer arcs along the $x$-coordinate of point $m$:
\begin{gather}
    r_{inner} =  \sqrt{|| arc_{inner} - O ||^2} \nonumber \\
    r_{outer} =  \sqrt{|| arc_{outer} - O ||^2}
\end{gather}
At this point, all relevant geometric information from the annulus sector has been processed. That is, the origin $O$, the angle $\theta$, and the inner and outer radii $[r_{inner}, r_{outer}]$.

\subsection{Scan line extraction and linearisation}

Most US machines, particularly clinical ones, do not provide access to the original radio frequency (RF) data or plane information, often restricting these features to users. For example, RF data can be used to accurately process the image in such a way that the phase information is preserved. A primary motivation for the proposed method is the acquisition of scan lines, which offer the closest approximation to the original raw signals in the absence of RF data. The extraction of scan lines provides several advantages, such as enabling explicit processing along scan lines such as in \cite{Karamalis2012UltrasoundCM}. Another application of this would be accurate data augmentation, which is particularly useful for enhancing data diversity in neural network training. This section demonstrates how the method proposed above can be applied to linearise a convex US plane. 

Using the estimated geometric parameters of the annulus sector, the US plane can be fanned over between the left and right radial boundaries to extract the start $[i_{inner}, j_{inner}]$ and end $[i_{outer}, j_{outer}]$ of the scan line vectors as:
\begin{gather}
    i_{inner} = r_{inner} \cos(\hat{\theta}) + O[1], j_{inner} = r_{inner} \sin(\hat{\theta}) + O[0], \nonumber \\
    i_{outer} = r_{outer} \cos(\hat{\theta}) + O[1], j_{outer} = r_{outer} \sin(\hat{\theta}) + O[0] 
\end{gather}
where, $\hat{\theta}$ is in the range $[90\degree - \theta/2, 90\degree + \theta/2]$ with $step$ which can be defined by dividing $\theta$ by the length of the outer arc. Knowing $\theta$ and $[r_{inner}, r_{outer}]$, spline interpolation can be performed to extract subpixel representations of the scan line vectors. The number of samples for interpolation and the length of the scan lines (i.e. the linear image height), can be calculated by first determining the area of the annulus sector $Area = \theta\times(r_{outer}^2 - r_{inner}^2)/2$, then the $Height=r_{outer} - r_{inner}$ and $Width=Area/Height$. This produces a $Ratio=Height/Width$ so that the number of samples for interpolation can be determined using $step \times Ratio$. This ratio is key to maintaining the correct resolution and can also be used to control the computation cost (for example down-sampling). The lineralised image $\Omega$ is estimated as:
\begin{gather}
         \Omega = [interpolate\left( I_{rgb}, \Gamma_{\hat{\theta}} \right), \forall \hat{\theta} \in \theta] \\
        \Gamma_{\hat{\theta}} = [[i_{inner}, j_{inner}], [i_{outer}, j_{outer}]] \nonumber \\         
\end{gather}
where, $\Gamma_{\hat{\theta}}$ is the set of all beginning and end points for all scan lines. The pseudocode of the full linearisation process is shown in Algorithm~\ref{eq:psuedo}.

\begin{algorithm}
\caption{Annulus Sector Linearisation}\label{alg:linearization}
\begin{algorithmic}[1]
\State \textbf{Input:} Convex ultrasound image \textbf{\(I_{\text{rgb}}\)}
\State \textbf{Output:} Linearized ultrasound image \textbf{\(\Omega\)}

\State \(I_g \gets \) Convert \(I_{\text{rgb}}\) to grayscale
\State \(I_b \gets \) Binarize \(I_g\) using z-score for foreground detection
\State \(I_c \gets \) Extract largest connected component from \(I_b\)

\State \(\hat{S} \gets \) Compute negative cumulative sum of top half of \(I_c\)
\State \(S \gets \) Apply moving average on \(\hat{S}\) to smooth
\State \(m \gets \) Calculate the average of the center of mass and the midpoint between minimas of \(S\)

\State \(arc_{inner}, arc_{outer} \gets \) Determine vertical intersections based on \(m\) in \(I_c\)
\State \(edges_{l}, edges_{r} \gets \) Determine radial boundary edges on \(I_c\)
\State \(\hat{RB}_{l}, \hat{RB}_{r} \gets \) Robust line fitting using RANSAC on \(edges_{l}, edges_{r}\)

\State \(residual_{l}, residual_{r} \gets \) Compute residuals for \(\hat{RB}_{l}, \hat{RB}_{r}\)
\State \(RB_{l}, RB_{r} \gets \) Apply symmetric correction based on geometric properties relative to \(top, bottom\)

\State \(O \gets \) Find intersection of \(RB_{l}\) and \(RB_{r}\) using Cramer's rule
\State \(r_{inner}, r_{outer}, \theta \gets \) Determine geometric parameters of annulus sector

\State \(\Omega \gets \) Perform spline interpolation using \(r_{inner}, r_{outer}, \theta\)
\end{algorithmic}
\label{eq:psuedo}
\end{algorithm}

\section{Experimental Setup and Results}

\begin{figure*}[ht]
\centering
\includegraphics[width=0.7\textwidth]{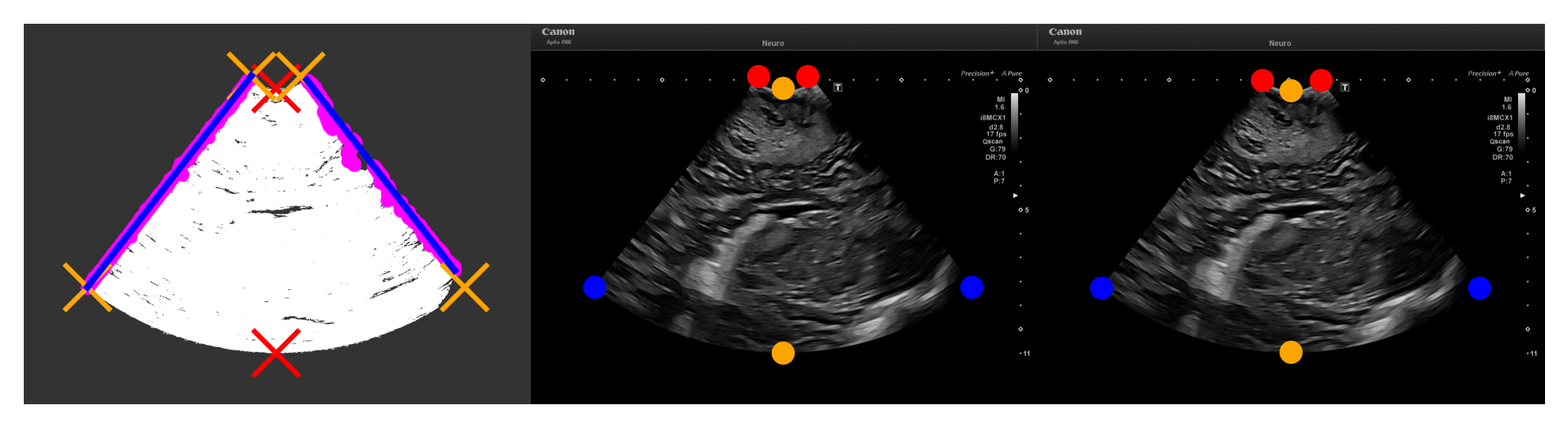} \\
\includegraphics[width=0.7\textwidth]{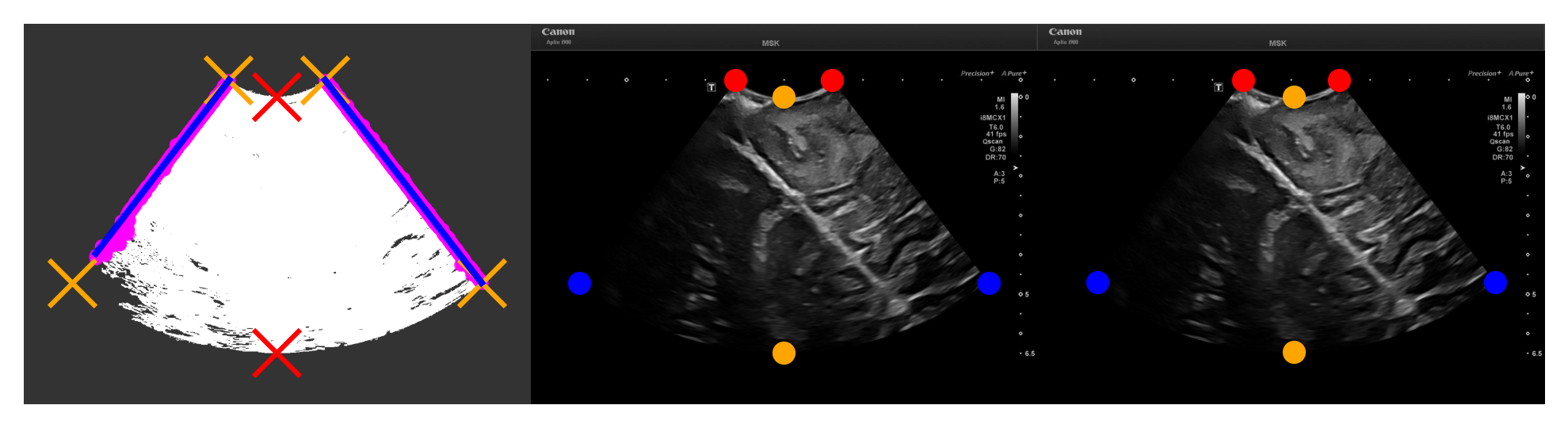}
\caption{Examples from the controlled experiment with the privaye data. The left images show the processing of the example images using the proposed method. The white plane is the output of the connected components process. The purple dots are the edge points of the mask, and the blue lines and orange crosses are the output of the RANSAC and the estimated radial boundaries. The middle and right images show the estimated and ground truth key points.}
\label{fig:robust}
\end{figure*}

\subsection{Data}
A controlled experiment, with minimal noise and corruption of the images, was carried out to validate the proposed approach on 30 images taken intraoperatively during brain surgery at Charing Cross Hospital, London, UK. A Canon i900 US machine (Canon Medical Systems, Japan) was used with an 8MHz i8MCX1 microconvex probe. The images are of size $960\times1280$. For the expansion of the evaluation to noisy public data, 50 selected images from POCUS \cite{pocus_dataset}, DUKE \cite{duke_dataset}, and CLUST \cite{clust_dataset} were processed and manually annotated. The images were annotated for all key points $[arc_{inner}, arc_{outer}, legs_{l,top}, legs_{l,bottom}, legs_{r,top}, \linebreak legs_{r,bottom}, O]$ that were used as our ground truth. 

\subsection{Key Point Detection Evaluation}

\subsubsection{Controlled Experiment}

In this section, we evaluate the accuracy of our method in extracting the annulus sector information from private US images. 
For this purpose, the Mean Square Error (MSE) (in pixels) is calculated for each key point separately, as:
\begin{equation}
    \text{MSE}^{kp} = \frac{1}{30} \sum_{f=1}^{30} \left( (j_{gt}^{(f,kp)} - j_{pred}^{(f,kp)})^2 + (i_{gt}^{(f,kp)} - i_{pred}^{(f,kp)})^2 \right)
\end{equation}
where, $kp$ is the key point index, $f$ is the image index, $[gt, pred]$ denote the ground truth and predicted estimates, and $i,j$ are the key point coordinates.

Across the 30 images, the following key point MSE were recorded: $arc_{inner}$ = 3.05 (std 2.38), $arc_{outer}$ = 3.24 (std 2.20), $legs_{l,top}$ = 4.41 (std 4.7), $legs_{l,bottom}$ = 4.50 (std 10.82), $legs_{r,top}$ = 5.90 (std 5.80), $legs_{r,bottom}$ = 6.51 (std 10.00), $O$ = 3.72 (std 2.83) (all in pixels). 

The error in the estimation of $\theta$ was also measured using the Mean Absolute Angular Difference (MAAD) (in degrees) as:
\begin{equation}
    \text{MAAD} = \frac{1}{30} \sum_{f=1}^{N} \left| \theta_{gt}^{f} - \theta_{pred}^{f} \right|
\end{equation}
The recorded MAAD for our dataset was 0.30$\degree$ (std 0.4$\degree$).

\subsubsection{Public, Noisy Data}

\begin{figure*}[ht]
\centering
\begin{minipage}[b]{0.45\textwidth}
    \centering
    \includegraphics[width=\textwidth]{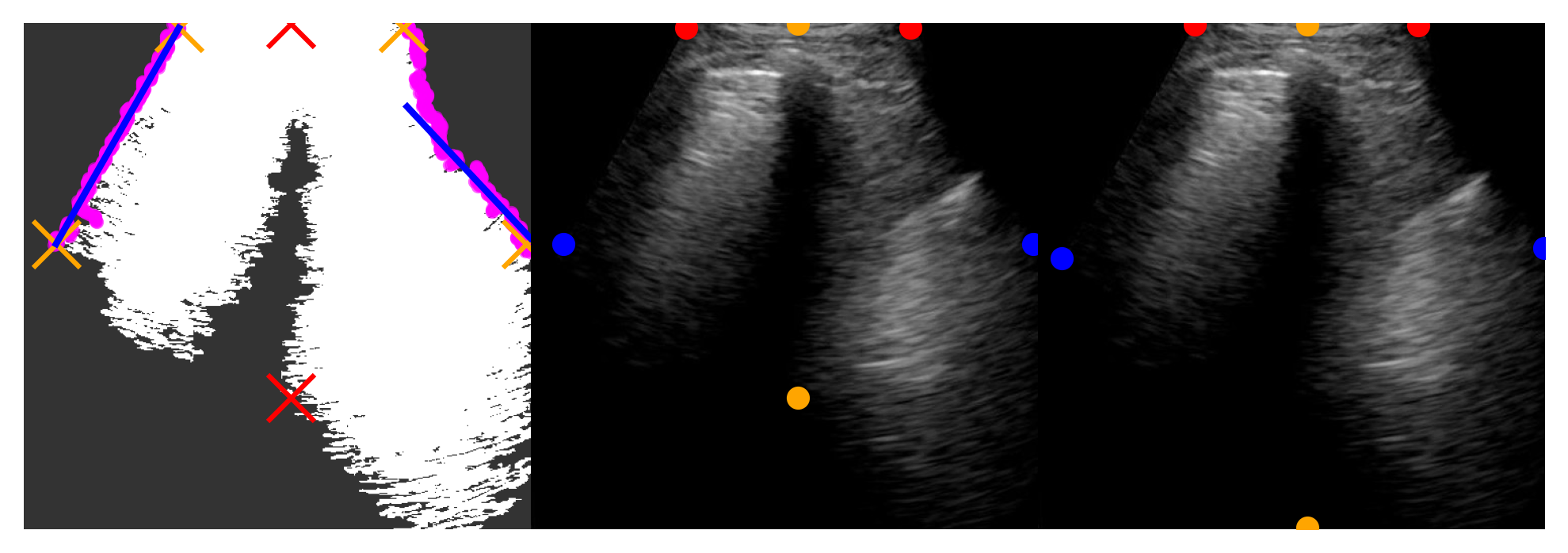}
    \parbox{1\textwidth}{\centering (a) POCUS dataset}
\end{minipage}
\hspace{0em}
\begin{minipage}[b]{0.35\textwidth}
    \centering
    \includegraphics[width=\textwidth]{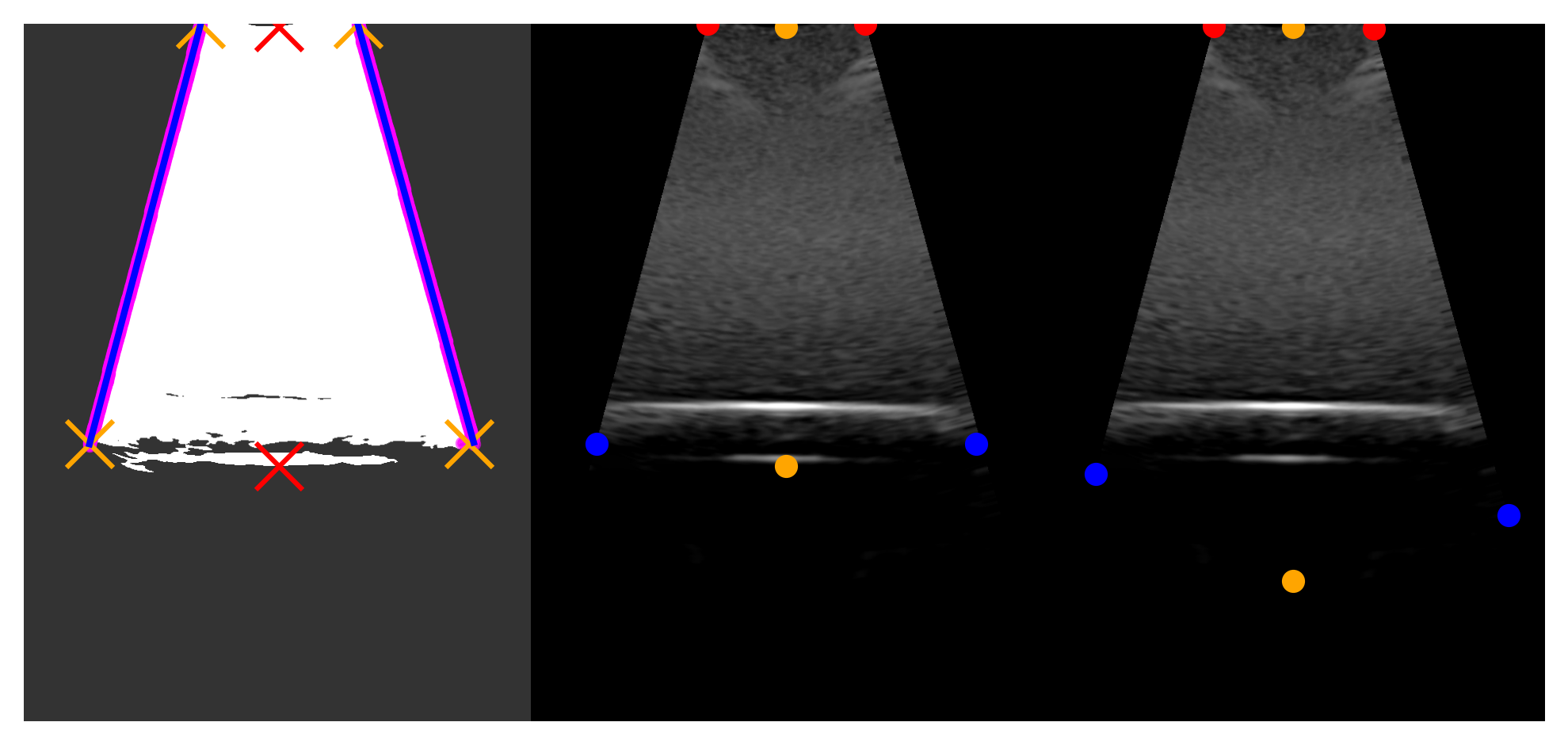}
    \parbox{0.8\textwidth}{\centering (b) DUKE dataset}
\end{minipage}

\vskip\baselineskip 

\begin{minipage}[b]{0.48\textwidth}
    \centering
    \includegraphics[width=\textwidth]{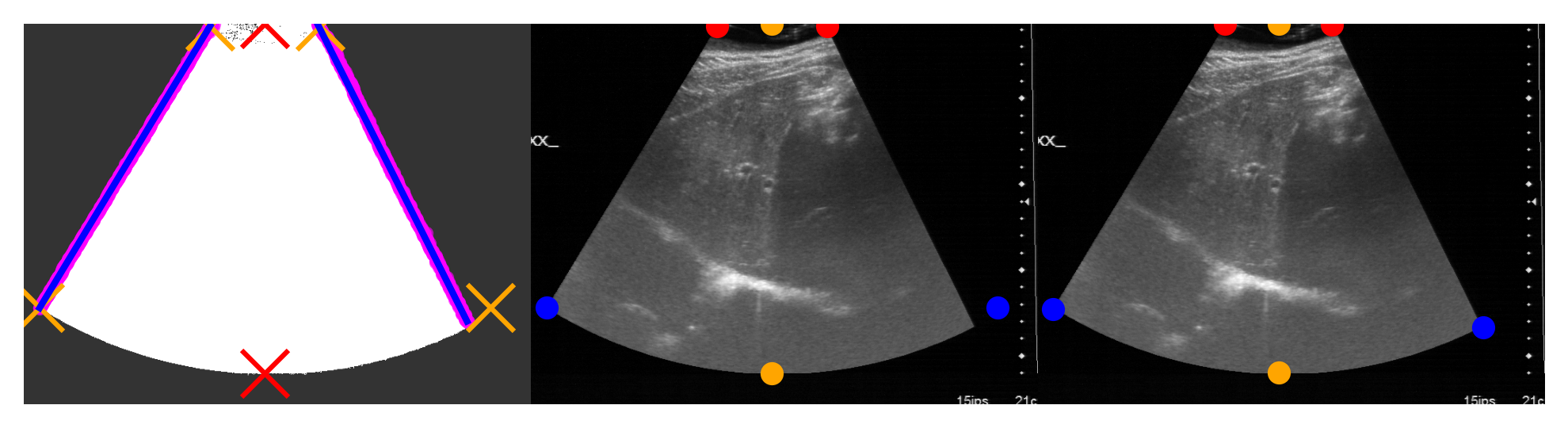}
    \parbox{0.8\textwidth}{\centering (c) CLUST dataset, video 05}
\end{minipage}
\hspace{0em}
\begin{minipage}[b]{0.49\textwidth}
    \centering
    \includegraphics[width=\textwidth]{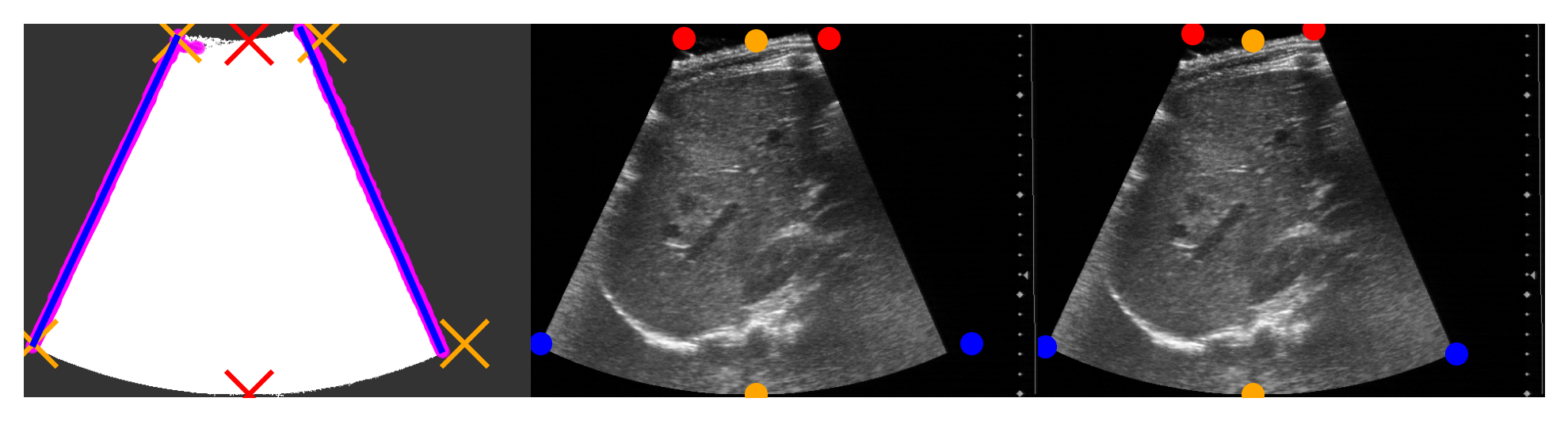}
    \parbox{0.8\textwidth}{\centering (d) CLUST dataset, video 09}
\end{minipage}

\caption{Example images from the public data experiment, one example from each public dataset used. The left image is the output of the method, the middle is the estimated key points and the right is the ground truth. All examples highlight the issue of cropping on the top of the US plane. The POCUS and DUKE examples show the issue of US plane corruption. Both CLUST examples suffer from slight rotation.}
\label{fig:noisy_data}
\end{figure*}

Public US images are rarely stored in a standardised format without corruption or alteration of the US plane. Often, images are cropped, rotated, warped, or noisy. In addition, they may be affected by other factors, such as capture from different devices and machines with varying settings. Examples of the data used in this evaluation are shown in Fig.~\ref{fig:noisy_data}. This degree of irregularity in the data makes it challenging to utilise different data sources together in a collective way. The method proposed in this paper is an approach to alleviate this issue by being able to extract the underlying geometric properties of any US image. To evaluate the performance of our method on diverse data, images from multiple public datasets were annotated. Specifically, 18 images were taken from POCUS \cite{pocus_dataset}, 17 from DUKE \cite{duke_dataset}, and 15 from CLUST \cite{clust_dataset}. The selected images were chosen for their variability in the issues mentioned above.  

There were two conditions used during data selection and annotation. Firstly, if the image was too challenging for human annotation, they would not be included in this test data. Secondly, the method is fundamentally designed for non-rotated or minimally rotated US-planes. Principally, this is because 2D US is not rotated in clinical practice. 
So, only non-rotated or slightly rotated images were chosen for annotation. This limitation mostly only concerns reconstructed 3D US - as 2D US (the primary concern of this paper and the more common US data) should not be stored rotated. In this case, slices are often poorly standardised and the US plane can vary significantly or even be absent due to cropping. The CLUST examples shown in Fig.~\ref{fig:noisy_data} are the cases where there is a slight rotation.

To handle images where the top of the US plane has been cropped - refer to all examples in Fig.~\ref{fig:noisy_data}. Normally, the crop will result in the US plane starting from outside the image plane. So, if the top rows of $I_c$ contain the binary mask (i.e. a vector of 1's at the top of the image), then the image is determined as not containing a concave profile. In this circumstance, rather than calculating the centre of mass, the middle point between the top left and right edges of the binary mask is used as the point $m$. The evaluation is carried out using the estimated and manually annotated geometric properties of the annulus sector: $O$, $\theta$, $arc_{inner}$, and $arc_{outer}$. The average shape of the POCUS images was $571 \times 571$, DUKE $1345 \times 960$, and CLUST $510 \times 686$. The results are presented in Tab.~\ref{tab:mse_maad}.



\begin{table}[h]
\centering
\large
\caption{MSE and MAAD results from the public data evaluation.}
\resizebox{\columnwidth}{!}{%
\begin{tabular}{|c|cc|cc|cc|cc|}
\hline
\textbf{Dataset} & \multicolumn{2}{c|}{\textbf{MSE ($O$)}} & \multicolumn{2}{c|}{\textbf{MSE ($arc_{inner}$)}} & \multicolumn{2}{c|}{\textbf{MSE ($arc_{outer}$)}} & \multicolumn{2}{c|}{\textbf{MAAD ($\degree$)}} \\ \hline
 & \textbf{Mean} & \textbf{Std} & \textbf{Mean} & \textbf{Std} & \textbf{Mean} & \textbf{Std} & \textbf{Mean} & \textbf{Std} \\ \hline
POCUS & 23.73 & 43.38 & 4.17 & 7.19 & 23.59 & 43.04 & 1.98 & 2.47 \\ \hline
DUKE & 14.70 & 8.76 & 2.19 & 1.57 & 22.75 & 54.78 & 0.80 & 0.44 \\ \hline
CLUST & 20.85 & 4.93 & 8.48 & 2.15 & 9.06 & 2.05 & 2.97 & 0.024 \\ \hline
\end{tabular}}
\label{tab:mse_maad}
\end{table}

\begin{figure*}[h]
\centering
\includegraphics[width=0.7\textwidth]{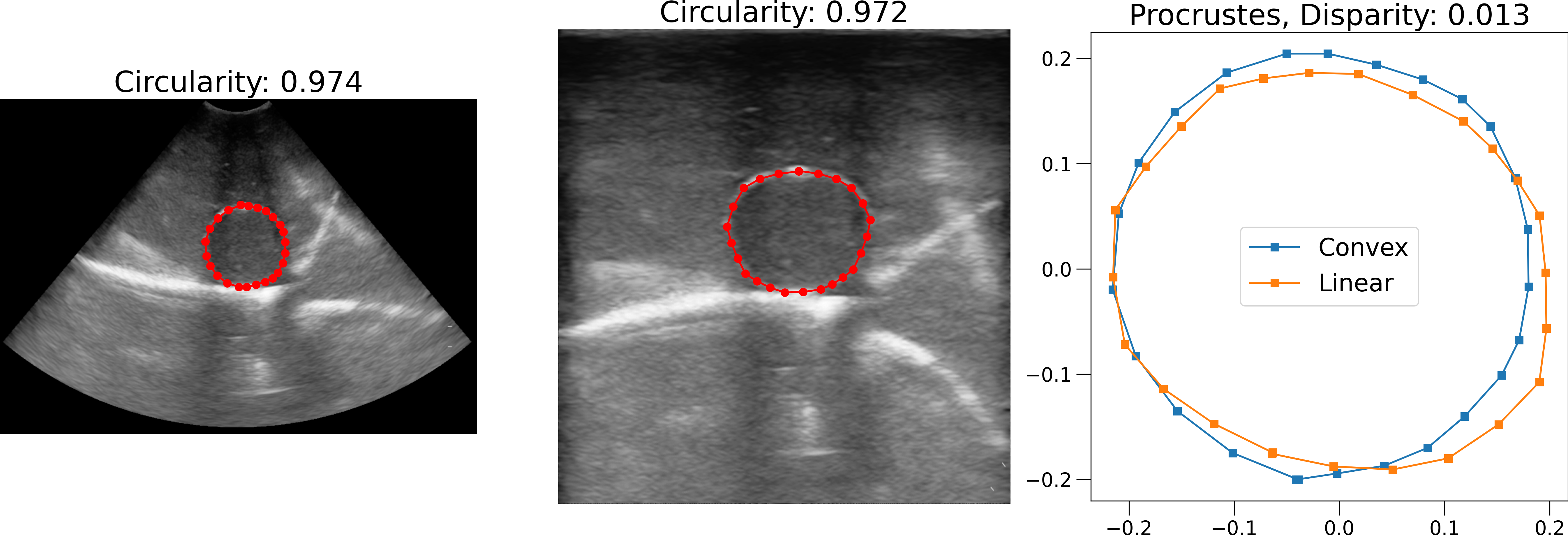}
\caption{Comparison of the circularity of the object in the original convex US plane compared to the linearised version. Quantitatively this comparison is measured using circularity and Proscrutes metrics.}
\label{fig:pingpong}
\end{figure*}

\begin{figure}[h]
\centering
\includegraphics[width=0.55\columnwidth]{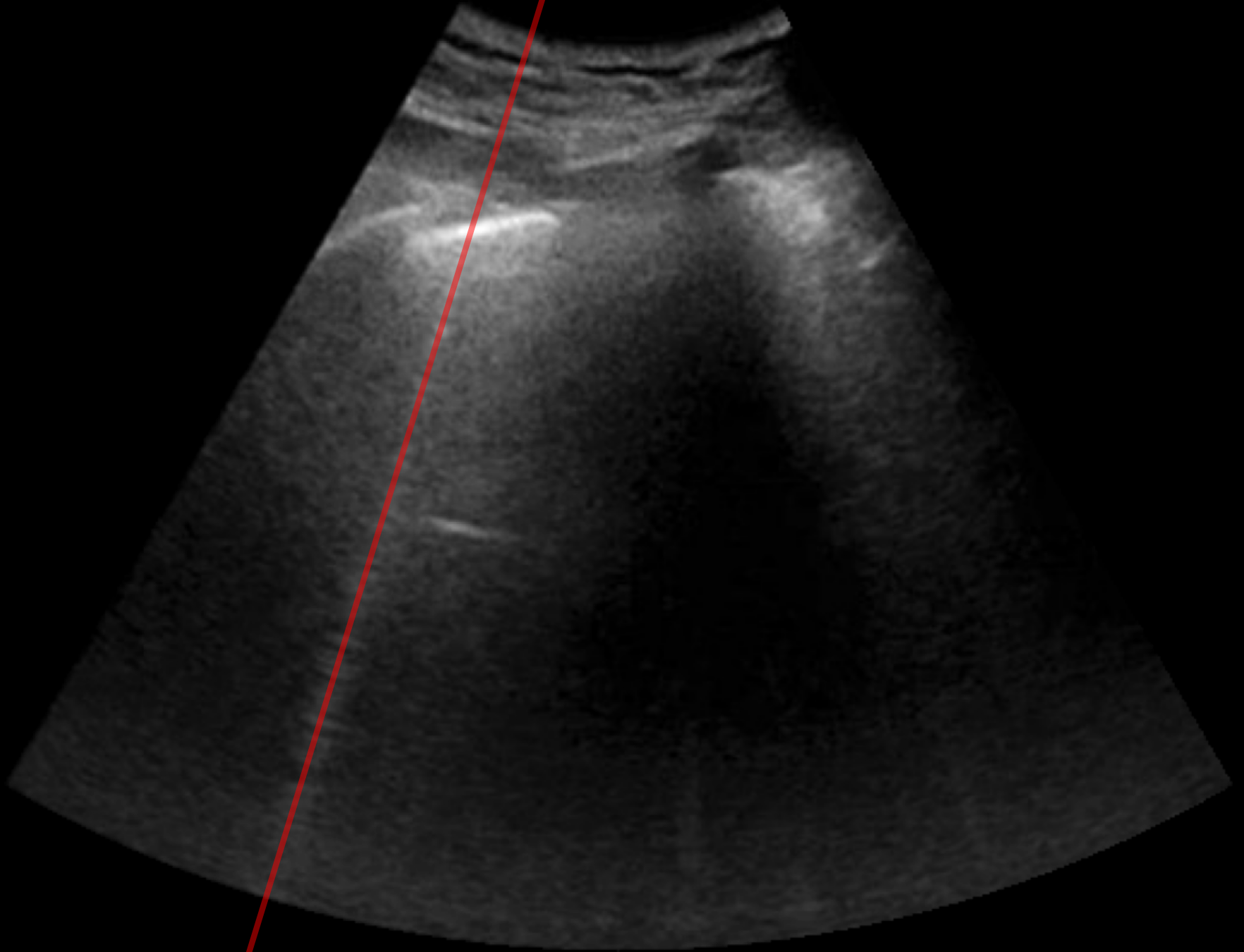}
\includegraphics[width=0.32\columnwidth]{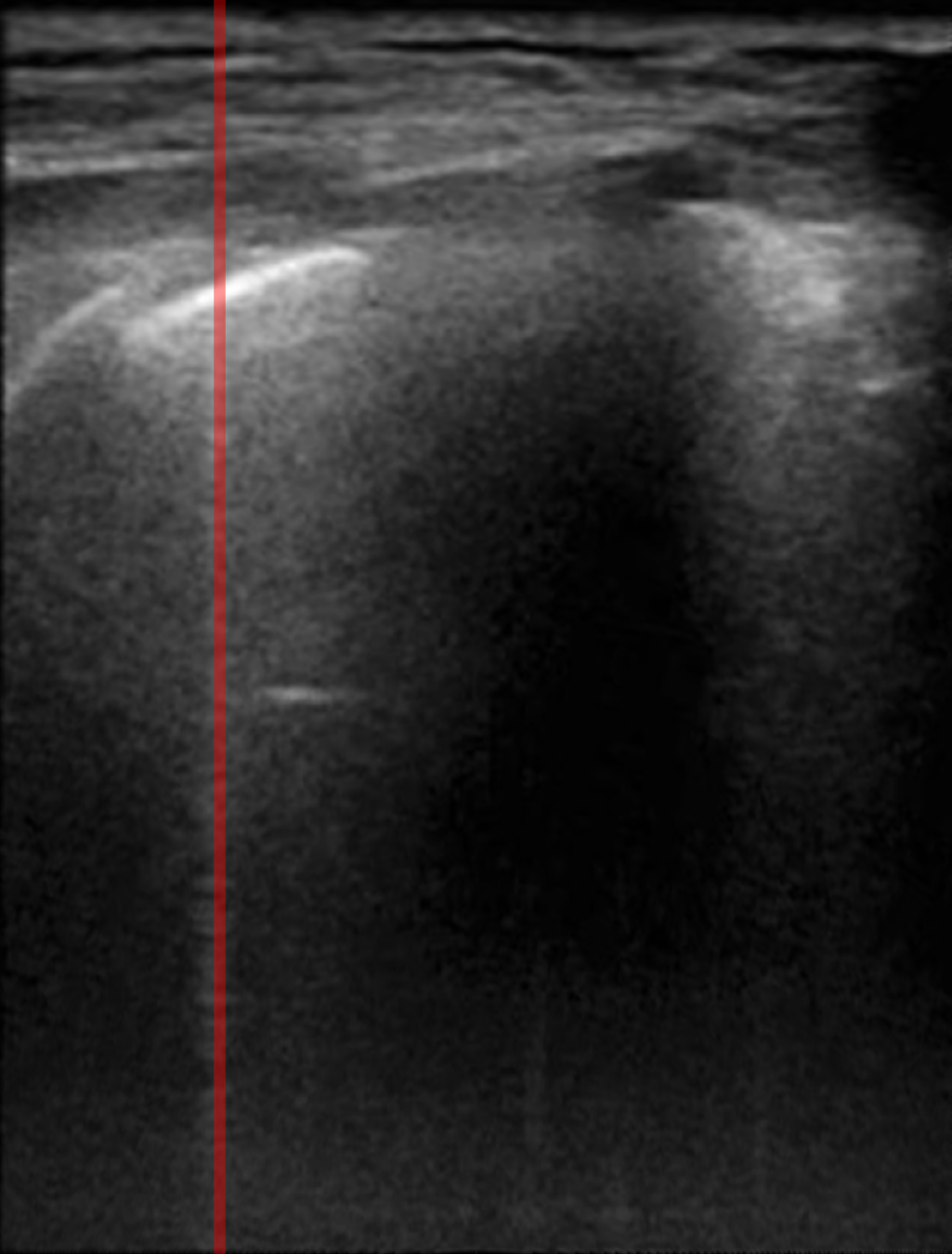}
\caption{Sample from POCUS dataset \cite{born2021accelerating} showing reverberation. The linearisation process correctly projects the reverberation as vertical. }
\label{fig:pocus1}
\end{figure}

\subsection{Linearisation Study}

\subsubsection{Deformation} Experiments were conducted to assess the degree of deformation that occurs after a convex plane is linearised using our method. The purpose of this evaluation is to validate that the proposed method can correctly maintain the geometric features after augmentation. First, data was collected of a ping pong ball placed in water, using a Zonare Z.One Ultrasound System. The circularity of the ball was measured in both the original convex and extracted linear US images.  For this purpose, in both images, the ping pong ball has been manually annotated and the circularity of both annotations is measured as:
\begin{equation}
    \text{Circularity} = (4\pi\text{Area})/\text{Perimeter}^2
\end{equation}
As it is shown in Fig.~\ref{fig:pingpong} (middle), the annotation on the extracted linear US image appears slightly stretched horizontally. However, this is somewhat expected as the linearisation will always deform the image to a certain degree. The quantitative results show that the circularity is almost equal in both images, approximately $0.97$. Procrustes analysis \cite{Gower1975GeneralizedPA} is used to further compare the annotations by normalising them with respect to each other through translation and then isomorphic scaling and rotation. The result of this analysis is shown in Fig.~\ref{fig:pingpong} (right). The recorded disparity, calculated as the sum of the squared pointwise differences is $0.013$. Additionally, it can be observed that the contact-related acoustic shadow, the black columns on the left and right of the circle, is correctly linearised, with both columns projected as vertical lines.



Reverberations are common artefacts that appear as hyperechoic lines along scan lines, caused by repeated reflections between interfaces, including pleural line artefacts such as B-lines. To examine this, an image from the POCUS dataset \cite{born2021accelerating} is evaluated. The left image of Fig.~\ref{fig:pocus1} shows a convex image with a red scan line which has been manually annotated by our clinical team as a reverberation. It can be assumed that the linearisation process should project reverberations as vertical columns. As can be observed, in the linearised image in Fig.~\ref{fig:pocus1} (right), the reverberation is projected vertically, again along a scan line. This result shows that the linearisation process has preserved the true properties of the image, by maintaining the physics of the reverberation, without deformation.




\subsubsection{Inversion} The aim of this section is to show that the linearisation process is not lossy and that the image augmentation process does not affect the underlying US information. The 30 intraoperative US images were linearised and then re-projected back to convex, using the extracted annulus sector geometry and $\hat{\theta}$. To measure the re-projection error, a direct MSE was calculated between the original and re-projected convex plane; across the 30 images the MSE was 0.0064 (std 0.0044) pixels.

\subsubsection{Radio-Frequency Data}

\begin{figure}[t]
\centering
\includegraphics[width=0.32\columnwidth, height=3.5cm]{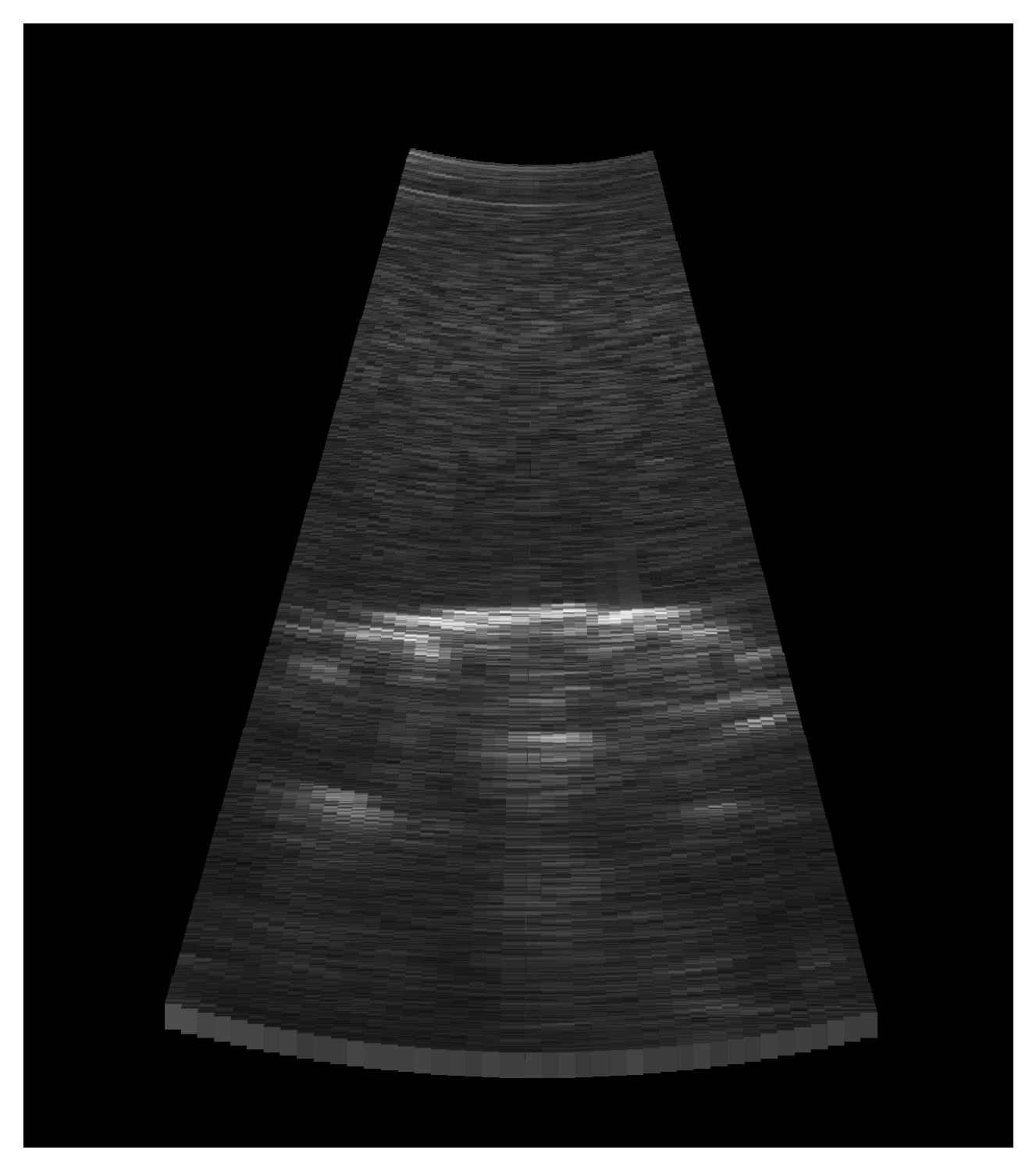}
\includegraphics[width=0.32\columnwidth, height=3.5cm]{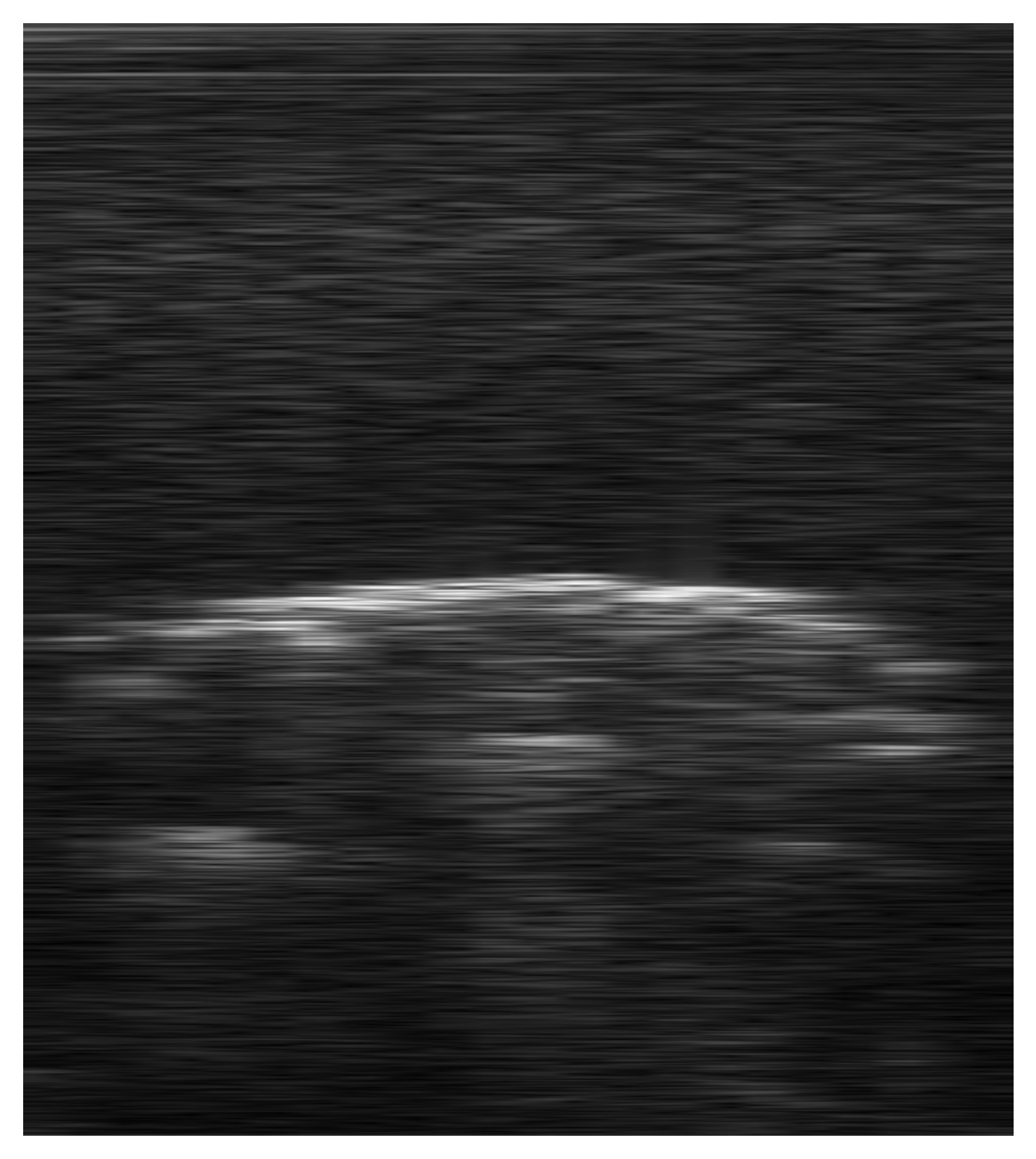}
\includegraphics[width=0.32\columnwidth, height=3.5cm]{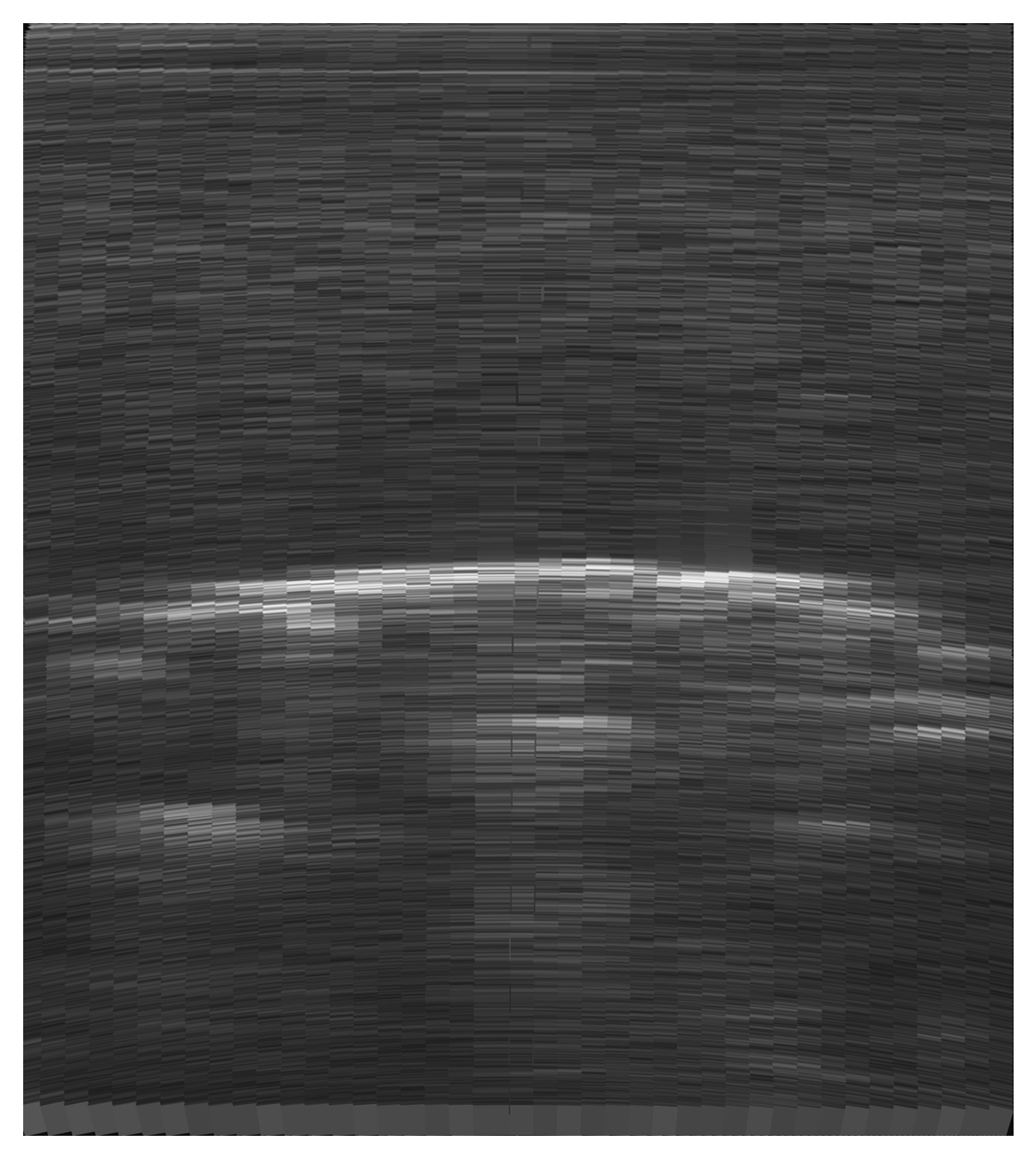}
\caption{Radio-frequency data validation. Left is the captured convex image. Middle is the ground truth linear image. Right is the linearised image estimated with our proposed method.}
\label{fig:rf}
\end{figure}

The evaluation of the proposed linearisation method is extended with the use of RF data, which enables quantitative analysis of the linear output. The aim is to generate ground truth linear images using RF data, which can then be used to compare with the  linearised images produced by the proposed method via structural similarity.

Access to raw RF data is not commonly available in clinical imaging systems, usually only in a limited number of research systems. 
B-mode US images provide a 2D visualisation of the RF data echo signals. A general representation of the detected echo $E^*(t)$ is:
\begin{gather}
E^*(t)=A(t)e^{j(2\pi ft+\varphi)}=  \nonumber \\
A(t)cos(2\pi ft +\varphi)+j_{rf}A(t)sin(2\pi ft +\varphi)=I+j_{rf}Q
\end{gather}
where, $A(t)$ represents the signal envelope, $f$ is the carrier frequency, $\varphi$ is the signal phase, and $I$ and $Q$ represent the in-phase and quadrature sinusoids of the signal, respectively. For convex and phased array probes, each echo has a corresponding $\theta_{rf}$, which relates to the orientation of the piezoelectric element. From the RF data, B-mode US images can be constructed by calculating the signal envelope, which then undergoes logarithmic compression and decimation. 

A US slice was captured using a phased array US probe (P5-1S15-A6, Telemed, Lithuania), a digital acquisition board (DAQ) (ArtUS, Telemed, Lithuania) and a phantom made of Agarose (2\% v/v, Sigma-Aldrich) with a Polyvinil alcohol (PVA) circular inclusion (4\% v/v, Sigma-Aldrich). Fig.~\ref{fig:rf} shows the captured US data (left and middle) and the output of the proposed method (right). 


Direct comparison of the intensities between the ground truth linear image from projected RF data and linearised images generated from convex B-mode is not feasible due to pixel discrepancy in the RF projection. Due to piezoelectric tilting $\theta_{rf}$, corresponding convex and projected linear B-mode images exhibit variations in resolution. These variations lead to pixel-level discrepancies, which arise from differences in scan line spacing, overlap, and angular coverage. Moreover, since the data were acquired using a phased array, the convex image is generated through a projection process, in which the inner null region is simulated by prepending a zero vector. To mitigate the impact of these pixel-level differences, the evaluation of image linearisation is performed at the structural level using the multi-scale structural similarity metric (MS-SSIM) \cite{Wang2003MultiscaleSS}.
\begin{equation}
\text{MS-SSIM} = [l_M] \prod_{g=1}^M [c_g][s_g]
\end{equation}
where, $l$ is the luminance, $c$ is the contrast and $s$ is the structural similarity between the two images, and $g$ denotes the scale index during downscaling where $M=5$ corresponds to the coarsest level. The result of this experiment was $\text{MS-SSIM}$ equal to $0.69$, which shows strong similarity between the ground truth linear and the proposed method's linear output. This discrepancy can be attributed towards the fundamental intensity differences on the pixel level between the convex and linear data. Through qualitative and quantitative analysis with ground truth, the results show that the proposed methodology is capable of accurately linearising convex US data.

\section{Conclusion}

This article proposes a new perspective and solution to alleviating the issue with online US data non-standardisation via harmonisation of data captured using different machines and data recording/saving methods. A method has been proposed to extract and represent arbitrary convex US planes, through estimation of its annulus sector parameters. This automatic estimation method is beneficial primarily because it enables standardisation of US datasets without needing additional manual annotation. Furthermore, through the estimation and processing of 2D convex planes using annulus sector properties, the scan lines can be extracted, and the US plane can be augmented, which could be used for neural network dataset supplementation. The proposed methodology is built on traditional image processing techniques and is designed to be robust to incomplete and corrupted presentations of US planes. Through evaluation of the accuracy of extracted key points, it can be concluded that our proposed method is capable of accurately finding the edges of the US plane and determining the parameters of the annulus sector. The evaluation is expanded to show that the annulus sector is capable of extracting the scan lines and projecting into a linear format, free of deformation.

\section*{Acknowledgment}
This research was supported by the UK Research and Innovation (UKRI) Centre for Doctoral Training in AI for Healthcare (EP/S023283/1), the Royal Society (URF$\setminus$R$\setminus$2 01014]), the NIHR Imperial Biomedical Research Centre and Canon Medical Systems. 

\bibliographystyle{plain}
\bibliography{bibs}

\end{document}